\pdfoutput=1

\documentclass[11pt]{article}
\usepackage[final]{colm2025_conference}
\usepackage{microtype}
\usepackage{hyperref}
\usepackage{url}
\usepackage{booktabs}

\usepackage{lineno}

\usepackage{amsmath}
\usepackage{times}
\usepackage{latexsym}
\usepackage{placeins}
\usepackage{wrapfig}

\usepackage[T1]{fontenc}
\usepackage{booktabs} 
\usepackage{longtable} 
\usepackage{tabularx}  
\usepackage{amsmath}   

\usepackage[utf8]{inputenc}
\usepackage{microtype}

\usepackage{inconsolata}

\usepackage{graphicx}
\usepackage{amsfonts}
\usepackage{float}
\usepackage{pgfplots}
\pgfplotsset{compat=1.18}

\definecolor{mylightgray}{RGB}{220,220,220}
\definecolor{mylightblue}{RGB}{202, 241, 202}
\newcommand{\topic}{\texttt{Prompt-Reverse Inconsistency}} 
\newcommand{\randomnessinconsistency}{\texttt{Randomness Inconsistency}} 
\newcommand{\paraphraseinconsistency}{\texttt{Paraphrase Inconsistency}} 
\newcommand{\topicshort}{\texttt{PRIN}} 
\newcommand{\direct}{\texttt{Direct Prompt}}  
\newcommand{\inverse}{\texttt{Reverse Prompt}}

\title{\topic: LLM Self-Inconsistency Beyond  Generative Randomness and Prompt Paraphrasing}

\author{Jihyun Janice Ahn \& Wenpeng Yin\\
Department of Computer Science \& Engineering\\
The Pennsylvania State University\\
University Park, PA 16802, USA \\
\texttt{\{jfa5672, wenpeng\}@psu.edu} \\
}

\begin{document}

\ifcolmsubmission
\linenumbers
\fi
\maketitle
\begin{abstract}

While the inconsistency of LLMs is not a novel topic, prior research has predominantly addressed two types of generative inconsistencies: i) \randomnessinconsistency: running the same LLM multiple trials, yielding varying responses; ii) \paraphraseinconsistency: paraphrased prompts result in different responses from the same LLM. \randomnessinconsistency~arises from the inherent randomness due to stochastic sampling in generative models, while \paraphraseinconsistency~is a consequence of the language modeling objectives, where paraphrased prompts alter the distribution of vocabulary logits. This research discovers \topic~(\topicshort), a new form of LLM self-inconsistency: given a question and a couple of LLM-generated answer candidates, the LLM often has conflicting responses when prompted ``\emph{Which are correct answers?}'' and ``\emph{Which are incorrect answers?}''.
\topicshort~poses a big concern as it undermines the credibility of LLM-as-a-judge, and suggests a challenge for LLMs to adhere to basic logical rules.
We conduct a series of experiments to investigate \topicshort, examining the extent of \topicshort~across different LLMs, methods to mitigate it, potential applications, and its relationship with \randomnessinconsistency~and \paraphraseinconsistency. As the first study to explore \topicshort, our findings offer valuable insights into the inner workings of LLMs and contribute to advancing trustworthy AI.
\end{abstract}

\section{Introduction}
Large language models (LLMs), despite their strong performance across various domains, often exhibit behaviors that diverge significantly from human reasoning. One well-known issue in their generative process is inconsistency. LLM inconsistency is widely recognized by researchers and users, and it can be mainly categorized into two types: 
\begin{itemize}
    \item \randomnessinconsistency: Even when given the same prompt, an LLM may generate different responses across multiple trials. This randomness arises due to factors such as sampling stochasticity, model non-determinism, and softmax and floating-point precision errors in the generation process.
    \item \paraphraseinconsistency: When a prompt is rephrased while maintaining the same meaning, the LLM’s response can still vary. This occurs because the reformulated prompt implicitly alters the probability distribution within the language model's objective function.
\end{itemize}

Beyond generative tasks, LLMs are increasingly used for discriminative reasoning—a crucial capability in applications such as AI-assisted judging, grading, and evaluation. However, a fundamental challenge arises: due to generative inconsistencies, LLMs often produce multiple, conflicting candidate answers for the same question. While the Self-Consistency method \citep{wangself} leverages majority voting to mitigate this issue, an alternative approach is to enhance LLMs' ability to self-select the correct answer from a given set of options. Unfortunately, LLMs also exhibit inconsistency in discriminative reasoning, which we term \topic\ (\topicshort).

\topicshort~arises when an LLM is tasked with evaluating multiple answer candidates and determining which are correct or incorrect. As shown in Table \ref{tab:example}, LLMs frequently provide conflicting judgments over the same set of answer choices. This inconsistency raises serious concerns regarding: \textbf{The reliability of LLM-as-a-judge}: Inconsistencies undermine their trustworthiness in high-stakes applications, such as automated grading, peer review, and legal analysis.
\textbf{Fundamental logical inconsistencies}: If LLMs frequently violate basic logical principles when making judgments, their utility as reasoning agents is severely limited.


\begin{wraptable}[15]{r}{0.6\textwidth}
    \centering
    \renewcommand{\arraystretch}{1.5}
    \begin{tabular}{p{0.55\textwidth}}
        \hline
        \textbf{Question}: if $n$ is an integer and $101\times n^2$ is less than or equal to 10,000, what is the greatest possible value of $n$? \\
        \textbf{Options}: A) 7, B) 8, C) 9, D) 10, E) 11\\
        \hline
        \textbf{Direct Prompt}: What are the \textcolor{red}{correct} answers? \\
        \textcolor{blue}{\textbf{GPT4}}: ``C''\\
        \textbf{Reverse Prompt}: What are the \textcolor{red}{incorrect} answers? \\
        \textcolor{blue}{\textbf{GPT4}}: ``C, D, E''\\
        \hline
    \end{tabular}
    \caption{\topicshort~example from GPT4 (March 28, 2025)}
    \label{tab:example}
\end{wraptable}

This paper conducts a systematic investigation of \topicshort~in both closed-source and open-source LLMs, including GPT-4 \citep{openai2023gpt4}, Llama-3-8B-Instruct, Llama-3.3-70B-Instruct \citep{Meta2024Llama3}, Falcon-40B \citep{almazrouei2023falconseriesopenlanguage}, Qwen 2.5-72B \citep{qwen2.5}, and Mixtral-8x22B-MoE \citep{jiang2024mixtralexperts}. We evaluate these models across three tasks—MATH \citep{hendrycks2021measuringmathematicalproblemsolving}, MathQA \citep{amini2019mathqainterpretablemathword}, and EquationInference \citep{lou2024aaar}—spanning various answer set sizes, context lengths, domains, and difficulty levels (from high school, college, to PhD-level problems). Specifically, we design experiments to answer the following six research questions: 
\textbf{$\mathcal{Q}_1$:} \textit{How do different LLMs exhibit \topicshort\ ?} 
\textbf{$\mathcal{Q}_2$:} \textit{How will model randomness and prompt paraphrasing affect \topicshort?}
\textbf{$\mathcal{Q}_3$:} \textit{How to mitigate \topicshort~in LLMs?} 
\textbf{$\mathcal{Q}_4$:} \textit{How does \topicshort~correlate with \randomnessinconsistency~and \paraphraseinconsistency?} 
\textbf{$\mathcal{Q}_5$:} \textit{How effective can \topicshort\ be leveraged to enhance task performance?}
\textbf{$\mathcal{Q}_6$:} \textit{How does  \topicshort~vary with different sizes of options?} 

Our findings reveal several key insights. First, \topicshort\ does not positively correlate with \randomnessinconsistency\ or \paraphraseinconsistency, as some LLMs with low levels of these inconsistencies, such as Llama-3 and Falcon, still exhibit high \topicshort. This suggests that while these models are more deterministic, they fail to maintain logical consistency between \direct\ and \inverse. Second, \topicshort~can be mitigated by incorporating explicit reasoning paths between the question and answer candidates before prompting the LLM to determine correctness. Additionally, providing explainable information for the negation in \inverse\ further reduces \topicshort. Third, combining both \direct\ and \inverse\ reasoning can outperform the Self-Consistency approach when selecting the final answer from a candidate pool. However, this improvement is primarily observed in top-performing models such as GPT-4 and GPT-4o, while weaker models like Llama-3 show little to no benefit, likely due to their weaker instruction-following capabilities.

Our contributions are threefold: i) this is the first study to discover and conduct an in-depth analysis of \topicshort; ii) we propose effective solutions to mitigate \topicshort~and explore ways to leverage it; and iii) our experimental findings not only enhance the understanding of this non-human-like discriminative behavior in LLMs but also raises critical concerns for applications where LLMs serve as judges or evaluators.

\section{Related Work}

 This section mainly discusses prior work studying \randomnessinconsistency~and \paraphraseinconsistency~particularly in LLMs.

\paragraph{\randomnessinconsistency.} \citet{bubeck2023sparksartificialgeneralintelligence}  brought attention to the issue of randomness-induced inconsistency of GPT4. Building on this, \cite{wang2025assessingconsistencyreproducibilityoutputs} conducted a comprehensive evaluation of LLM consistency and reproducibility in finance and accounting tasks, highlighting the practical consequences of such variability. Similarly, \citet{DBLP04667} systematically examined LLM stability by repeatedly running identical inputs, revealing up to 10\% variation in output accuracy even under deterministic settings. Beyond quantitative assessment, \cite{DBLPLeeS0KCAK24} explored how these inconsistencies affect users, finding that while they may reduce perceived AI reliability, they can also enhance user comprehension by presenting diverse perspectives. To address these issues, \cite{wan2025reasoningawareselfconsistencyleveraging} proposed a sufficiency scoring method that evaluates both local and global consistency in LLM responses, offering a framework to analyze and mitigate instability driven by randomness.

\paragraph{\paraphraseinconsistency.}  \cite{elazar2021measuringimprovingconsistencypretrained} explored factual consistency across different query patterns and showed that while some paraphrase forms reliably extracted factual knowledge, others failed, revealing the model’s paraphrasing sensitivity. Similarly, \cite{ye2023assessinghiddenrisksllms} investigated this phenomenon in ChatGPT and found that response accuracy fluctuated by 3.2\% across paraphrased prompts, highlighting the influence of grammatical and stylistic variations on model behavior. 
Supporting this line of work, \cite{jang2023consistencyanalysischatgpt} documented cases of self-contradictions in ChatGPT and GPT-4 when exposed to paraphrased questions, confirming that even minor linguistic variations can lead to semantic inconsistencies. \cite{gu2023robustnesslearningtaskinstructions} extended this observation to instruction-driven tasks, demonstrating that LLMs often falter when task instructions vary in form, length, or abstraction, which further complicates generalization across paraphrased input formats.
In addition to task-specific studies, \cite{liu2024trustworthyllmssurveyguideline} provided a broader survey on LLM trustworthiness, in which they discussed inconsistency as a core reliability issue and emphasized the need for robust solutions to mitigate its effects. Complementing these discussions, recent work has introduced quantitative approaches to analyze and address paraphrase sensitivity. For example, \citet{DBLP12334} proposed metrics that measure how minor prompt variations influence LLM predictions in text classification tasks, offering a fine-grained assessment of response stability.
Further, \cite{ghazarian2024assessment} examined structural variations in semantically equivalent prompts and found notable inconsistency in LLM-based evaluations. They proposed an in-context learning strategy with demonstrations to improve robustness against paraphrasing. Finally, \cite{DBLPYoungBO24} tackled a related issue of order dependency in prompts and introduced Set-Based Prompting, a method designed to ensure consistent model behavior regardless of the sequence of sub-inputs, offering a new angle on mitigating paraphrase-driven inconsistencies.

\paragraph{Our Work}  
 is the first to explore \topic, not only analyzing this LLM behavior but, more importantly, proposing simple yet effective methods to mitigate the issue. Additionally, we examine its connection to \randomnessinconsistency~and \paraphraseinconsistency, as well as ways to leverage this inconsistency for improved model reliability.

\section{\topic}\label{sec:inconsistency}

\paragraph{Problem formulation.} Assume the prompt $p$, and LLM $\mathcal{M}$, multiple trials of $\mathcal{M}(p)$ leads to $n$ distinct answer candidates  $A=\{a_1, a_2, \cdots, a_n\}$ with each candidate $a_i$ derived through a Chain-of-Thought \citep{wei2022chain} reasoning path $r_i$. The task now is to figure out the correct answer from the pool $\{a_1, a_2, \cdots, a_n\}$ by querying $\mathcal{M}$ again. In this work, we study $\mathcal{M}$'s discriminative behavior through the following two prompts.

\paragraph{\direct.} Given the prompt $p$, answer options $\{a_1, a_2, \cdots, a_n\}$, it asks the correct one directly, e.g., 

\begin{quote}
\colorbox{gray!20}{\parbox{\linewidth}{%
Given this question [\emph{problem description}] and its answer options:  ``$a_1$'',  ``$a_2$'',  $\cdots$, ``$a_n$'', please output the \textcolor{red}{correct} ones.}}
\end{quote}

\paragraph{\inverse.}
Conversely, the models determine the incorrect choices as follows:

\begin{quote}
\colorbox{gray!20}{\parbox{\linewidth}{%
Given this question [\emph{problem description}] and its answer options:  ``$a_1$'',  ``$a_2$'',  $\cdots$, ``$a_n$'', please output the \textcolor{red}{incorrect} ones.}}
\end{quote}

\paragraph{\topicshort~Metric (The lower, the better).} Given the entire answer pool $A$, assuming \direct~return answer set $A_{direct}$ and \inverse~returns $A_{reverse}$, our metric is defined based on this rule: \textbf{if the correct answer sets derived by both prompts are the same, then no \topicshort}.

Therefore, first, the correct answer set by \direct~is $A_{direct}$. Then the correct answer set by \inverse~is $A\setminus A_{reverse}$. Then we compute the similarity of the two versions of correct answer sets through F1:
\begin{equation}
    s = F1(A_{direct}, A\setminus A_{reverse})
\end{equation}
then the \topicshort~score is:
\begin{equation}
    \topicshort=1.0-s
\end{equation}

\textbf{Question: Why not define the \topicshort~score as the similarity between $A_{direct}$ and $A_{reverse}$, i.e., $F1(A_{direct}, A_{reverse})$?}  

Intuitively, if $A_{direct}$ and $A_{reverse}$ are completely complementary (e.g., $A_{direct} = \{a_1, a_2\}$ and $A_{reverse} = \{a_3, a_4, \dots, a_n\}$), it implies no \topicshort. However, in practice, the union of their answers may not cover the entire answer pool. For instance, if $A_{direct} = \{a_1, a_2\}$ but $A_{reverse} = \{a_{n-2}, a_n\}$, using F1 as a measure would result in $F1(A_{direct}, A_{reverse}) = 0.0$, incorrectly indicating no \topicshort. This is problematic because $A_{reverse} = \{a_{n-2}, a_n\}$ suggests that the \inverse~prompt considers \{$a_1, a_2, \dots, a_{n-3}, a_{n-1}$\} as correct, which clearly reflects inconsistency to $A_{direct} = \{a_1, a_2\}$.

\section{Experiments}

\paragraph{Datasets.} We select the following three representative datasets: \textbf{Math} \citep{hendrycks2021measuringmathematicalproblemsolving}: This dataset consists of Math Word Problems, where each question $p_i$ is accompanied by the correct answer $a_i$ and a corresponding Chain-of-Thought reasoning path $r_i$. \textbf{MathQA} \citep{amini2019mathqainterpretablemathword}: A multiple-choice math dataset in which each Math Word Problem $p$ is presented with five answer choices, only one of which is correct. Unlike the Math dataset, reasoning paths are not provided. \textbf{EquInfer} \citep{lou2024aaar}: Designed to simulate the paper review process, this dataset evaluates equation correctness within a given context in a scientific paper. Each instance contains four equation candidates, with only one being correct, along with the surrounding paper context before and after the equation. 

This dataset selection demonstrates that the \topic~problem arises in both generative tasks (e.g., MATH) and discriminative tasks (e.g., MathQA and EquInfer), highlighting its broader implications. Table \ref{tab:benchmarks} summarizes key properties of these datasets.

\begin{table}[h] 
    \centering
    \resizebox{\textwidth}{!}{
        \renewcommand{\arraystretch}{1.2}
    \begin{tabular}{l|p{3.5cm}|c|c|c|c}
        \hline
         & \multicolumn{1}{c|}{\textbf{Format}} & \textbf{Size} & \textbf{Options} & \textbf{Context} & \textbf{Complexity} \\ 
        \hline
        \textbf{MATH} &  $p$ & 5,000 & None & Short  & High School \\ 
        \hline
        \textbf{MathQA} & ($p$; \{$a_1, a_2, a_3, a_4, a_5$\}) & 2,985 & 5 & Medium  & College \\ 
        \hline
        \textbf{EquInfer} & ($p$; \{$a_1, a_2, a_3, a_4$\}) & 1,049 & 4 & Long  & Ph.D.\\ 
        \hline
    \end{tabular}
    }
    \caption{Summary of three datasets (MATH, MathQA, and EquInfer).}
    \label{tab:benchmarks}
\end{table}

\paragraph{LLMs.} 
The experiment utilizes a combination of one closed-source model, GPT-4\footnote{Due to budget and administrative approval constraints, we cannot report on other closed-source LLMs.}, alongside five open-source models: GPT-4  \citep{openai2023gpt4} Llama-3-8B-Instruct (Llama3) and Llama-3.3-70B-Instruct (Llama3.3) \citep{Meta2024Llama3}, Falcon-40B (Falcon) \citep{almazrouei2023falconseriesopenlanguage}, Qwen 2.5-72B (Qwen2.5) \citep{qwen2.5}, and Mixtral-8x22B-MoE (Mixtral) \citep{jiang2024mixtralexperts}. 

\paragraph{Setting.} 
To prepare for the core experiments, we generate answer options for the MATH dataset, as they are absent in the original benchmark. GPT-4 solves each problem multiple times to produce five distinct answer choices, each with a Chain-of-Thought path, ensuring a uniform five-option format for the main experiment.  As the EquInfer has text of both sides as a problem description and 4 options, due to the token limitation for LLMs, 200 words for each side of the context are given to LLMs according to the suggestion by \cite{lou2024aaar}.


\subsection{\texorpdfstring{$\mathcal{Q}_1$}{Q1}: How do different LLMs exhibit \topicshort?}

Table \ref{tab:q1_table}  reveals a consistent pattern of \topicshort~across all evaluated LLMs, highlighting it as a fundamental and unresolved challenge. GPT-4 exhibits the lowest \topicshort~scores across all benchmarks, indicating that its superior instruction-following and reasoning abilities help mitigate, but not eliminate, inconsistency, as its \topicshort~still hovers around 40\%. Open-source models, including Qwen2.5, Mixtral, Falcon, Llama3, and Llama3.3, show significantly higher \topicshort~values, often exceeding 60\% on MathQA and EquInfer, suggesting that their reasoning abilities are particularly vulnerable when faced with reversed prompts. Interestingly, Qwen2.5 consistently outperforms other open-source models, possibly due to stronger instruction tuning, positioning it as the most robust among its peers. Moreover, the comparison between Llama3 and Llama3.3 shows that, despite architectural similarities, Llama3.3 reduces \topicshort~on MathQA and EquInfer but unexpectedly worsens on MATH, hinting that \topicshort~may be sensitive to domain-specific generalization. The consistently higher \topicshort~on EquInfer across models suggests that this dataset poses unique challenges, likely due to its demand for nuanced reasoning under prompt reversals. Overall, the results indicate that while advanced models like GPT-4 alleviate \topicshort~to some extent, significant inconsistency persists across all models, pointing to \topicshort~as a critical barrier to trustworthy reasoning in LLMs.

\begin{table}[t]
    \centering
    \begin{tabular}{l|ccc|c}
         & MATH & MathQA & EquInfer & Mean \\\hline   
        GPT4 & \textbf{38.69} & \textbf{38.60} & \textbf{42.65} & \textbf{39.98}\\
        Qwen2.5 & 58.41 & 51.31 & 70.82 & 60.18 \\
        Mixtral & 67.77 & 63.58 & 74.83 & 68.73\\
        Llama3.3 & 80.96 & 61.17 & 76.62 & 72.92\\
        Falcon & 71.79 & 68.69 & 83.42 & 74.63\\
        Llama3 & 74.10 & 84.08 & 80.46 & 79.55\\
    \end{tabular}
    \caption{\topicshort~scores for all LLMs (answers to $\mathcal{Q}_1$).} 
    \label{tab:q1_table}
\end{table}

\subsection{\texorpdfstring{$\mathcal{Q}_2$}{Q2}: How will model randomness and prompt paraphrasing affect \topicshort?} 

In this subsection, we explore if \direct~and \inverse~still show inconsistency even if we i) paraphrase them (apply \paraphraseinconsistency), or ii) run them multiple times (apply \randomnessinconsistency).

First, as Table \ref{tab:prompt_versions} shows, we paraphrase \direct~and \inverse~introduced in Section \ref{sec:inconsistency} into two new versions.
\begin{table}[!htbp]
    \centering
    \begin{tabular}{p{1.5cm}|p{3cm}|p{3cm}|p{3cm}}
         & \multicolumn{1}{c|}{original} & \multicolumn{1}{c|}{paraphrased prompt 1}& \multicolumn{1}{c}{paraphrased prompt 2} \\  
         & \multicolumn{1}{c|}{(v0)} & \multicolumn{1}{c|}{(v1)} &  \multicolumn{1}{c}{(v2)}\\\hline   
        \direct\  & Please output the \textbf{correct} ones. & Please output the \textbf{right} ones. & Please output the \textbf{appropriate} ones. \\\hline  
        \inverse\ \  & Please output the \textbf{incorrect} ones. & Please output the \textbf{wrong} ones. & Please output the \textbf{inappropriate} ones. \\
    \end{tabular}
    \caption{Paraphrased \direct~and \inverse.}
    \label{tab:prompt_versions}
    \vspace{-20pt}
\end{table}

\paragraph{Results of applying paraphrasing:} Table \ref{tab:paraphrase} presents the effects of prompt paraphrasing on \topicshort~scores. The \textbf{variations in scores} indicate the presence of \paraphraseinconsistency, demonstrating that LLMs' responses are influenced by how prompts are phrased. However, the \textbf{changes are relatively minor}, suggesting that \topicshort~remains largely stable across paraphrased inputs. This implies that while LLMs are somewhat sensitive to different prompt formulations, their \topicshort~follows a systematic pattern rather than being highly volatile due to rewording alone.

\begin{table}[h]
    \centering
    \begin{tabular}{l|c|c|c|c|c|c|c|c|c}
         & \multicolumn{3}{c|}{MATH} & \multicolumn{3}{c|}{MathQA} & \multicolumn{3}{c}{EquInfer} \\
         & v0 & v1 & v2  & v0 & v1 & v2  & v0 & v1 & v2  \\\hline  
         
        GPT4 & \textbf{38.69} & \textbf{38.81} & \textbf{37.22}  & \textbf{38.60} & \textbf{39.48} & \textbf{38.14}  & \textbf{42.65} & \textbf{41.89} & \textbf{48.34} \\
        
        Qwen2.5 & 58.41 & 56.63 & 60.02  & 51.31 & 50.55 & 58.11  & 70.82 & 71.76 & 69.73  \\
        
        Mixtral & 67.77 & 67.29 & 69.10  & 63.58 & 68.04 & 73.31  & 74.83 & 78.32 & 74.47  \\
        
        Llama3.3 & 80.96 & 81.40 & 78.39  & 61.17 & 59.82 & 59.17  & 76.62 & 73.51 & 76.28  \\
        
        Falcon & 71.79 & 72.06 & 71.74  & 68.69 & 71.74 & 73.68  & 83.42 & 86.22 & 81.89  \\
        
        Llama3 & 74.10 & 73.69 & 72.91  & 84.08 & 83.40 & 81.88  & 80.46 & 80.76 & 80.86  \\
    \end{tabular}
    \caption{Effect of prompt paraphrasing on inconsistency across tasks.}
    \label{tab:paraphrase}
\end{table}

Next, we conduct five repeated runs of the original \direct~and \inverse~(i.e., v0) prompts for each LLM to assess the consistency of their results. Table~\ref{tab:trials} presents the mean and standard deviation of \topicshort, confirming that \topicshort~remains stable with only minor fluctuations across runs. Together, Tables \ref{tab:paraphrase}-\ref{tab:trials} demonstrate that \topicshort~is not an artifact of a particular choice of \direct~and \inverse~prompts but rather a systematic issue that persists across a wide range of LLMs.
\begin{table*}[h]
    \centering
    \begin{tabular}{l|c|c|c}
         & MATH & MathQA & EquInfer \\\hline
        GPT4 & 38.66$\pm$0.29 & 39.54$\pm$0.17 &  42.81$\pm$0.73\\
        
        Qwen2.5 & 58.67$\pm$0.58 &  52.64$\pm$0.41 &  69.75$\pm$0.44 \\
        
        Mixtral &  67.74$\pm$0.42 &  67.08$\pm$0.62 &  74.42$\pm$0.62 \\
        
        Llama3.3 &  80.81$\pm$0.32 &  59.35$\pm$0.64 &  76.36$\pm$0.68 \\
        
        Falcon &  71.01$\pm$0.15 &  68.94$\pm$0.58 &  81.16$\pm$0.50 \\
        
        Llama3 &  74.56$\pm$0.35 &  83.93$\pm$0.51 &  80.70$\pm$0.77 \\
    \end{tabular}
    \caption{\topicshort~scores when we run \direct~and \inverse~five times. }
    \label{tab:trials}
\end{table*}

\begin{figure}[ht]   
    \centering
    \includegraphics[width=0.7\textwidth]{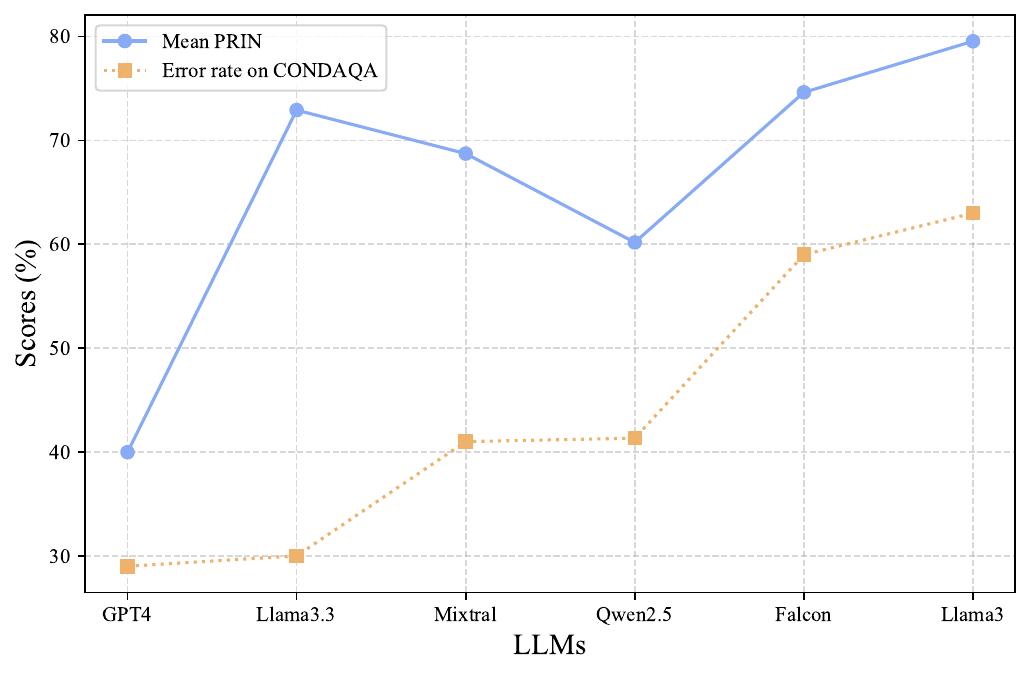}
    \caption{Mean \topicshort~on three main benchmarks vs. error rates on CONDAQA} 
    \label{fig:condaqa_vs_prin}
\end{figure}

\paragraph{Question: Table~\ref{tab:prompt_versions} suggests that \inverse~prompts often involve negation. How well does LLMs' \topicshort~align with their performance on a negation-specific task?} To investigate this, we evaluate the LLMs on the negation-focused dataset CONDAQA~\citep{ravichander2022condaqa} and compare their \topicshort~scores (``Mean'' column in Table~\ref{tab:q1_table}) with their error rates on CONDAQA. As shown in Figure~\ref{fig:condaqa_vs_prin}, the two measures exhibit a strong alignment, with a Pearson correlation coefficient of 0.67. This result confirms that the core challenge captured by \topicshort~is closely related to the models' difficulty in handling negation.


\subsection{\texorpdfstring{$\mathcal{Q}_3$}{Q3}: How to mitigate \topicshort~in LLMs?}

\paragraph{Our Approach:}  
To ensure the broad applicability of our investigation into \topicshort, the aforementioned experiments were conducted with each query $p$ paired with a pool of answer candidates $A = \{a_1, a_2, \dots, a_n\}$. However, in real-world scenarios, humans may better distinguish between \direct~and \inverse~when they understand how each answer candidate was derived. Motivated by this, our first approach incorporates CoT reasoning paths $r_i$ for each answer candidate $a_i$, allowing for a more informed evaluation of \topicshort. We refer to it as ``\textbf{w/ CoT}''.

\begin{wrapfigure}[17]{r}{0.55\textwidth}
  \begin{center}
  \vspace{-0.1in}
    \includegraphics[width=0.5\textwidth]{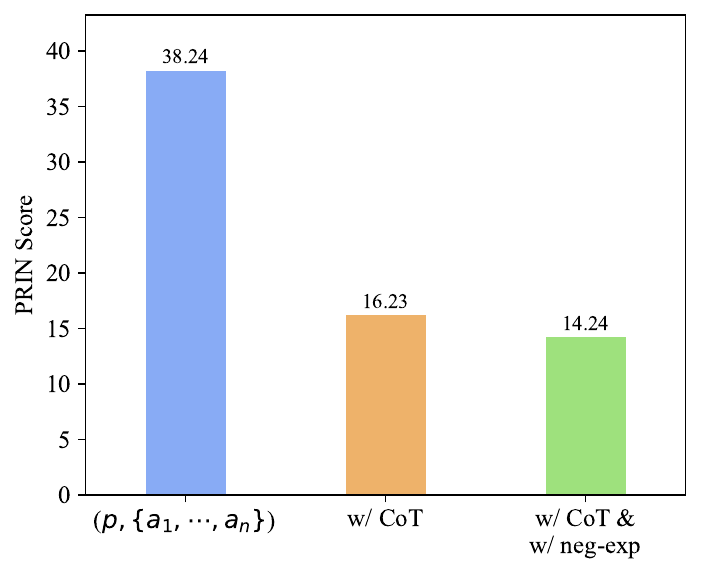}
  \end{center}
  \vspace{-5mm}
  \caption{\topicshort~score of GPT4 on MATH benchmark with different mitigation approaches. }
  \label{fig:mitigation}
\end{wrapfigure}
Our second approach is inspired by the observations (to $\mathcal{Q}_2$) that discrepancies between \inverse~and \direct~may arise due to the LLMs' difficulty in processing negation in \inverse. To address this, we enhance the clarity of negation terms by explicitly explaining their meaning within \inverse. One simple sentence such as \textit{``please recall that `incorrect options' are simply the options different from the correct ones."} was added in the end of the \inverse. The same evaluation metric is then applied to assess the impact of this intervention. We refer to this approach as ``\textbf{w/ neg-exp}''.

\paragraph{Results:} Figure \ref{fig:mitigation} demonstrates the effectiveness of our \topicshort~mitigation approaches. Incorporating detailed reasoning ensures models make informed decisions based on deeper understanding rather than rote selections. The empirical evidence highlights the importance of contextual reasoning in improving AI comprehension and reducing errors. Moreover, reinforcing models with explicit explanations about negation in inverse tasks further decreases \topicshort. Providing clarifications, especially regarding negation terminology, aids in reducing confusion and logical pitfalls, establishing an effective strategy for error reduction.


\subsection{\texorpdfstring{$\mathcal{Q}_4$}{Q4}: How does \topicshort~correlate with \randomnessinconsistency~and \paraphraseinconsistency?}

\paragraph{Setup.}  
To investigate $\mathcal{Q}_4$, we quantitatively assess \randomnessinconsistency~and \paraphraseinconsistency~across various LLMs.  

For \randomnessinconsistency, we run \direct~five times and count the number of distinct answers $k$, computing the score as $k/5$.  For \paraphraseinconsistency, we use five paraphrased versions of \direct, recording $k$ distinct answers and defining the score as $k/5$. Since we already have three paraphrased versions (v0, v1, v2) from $\mathcal{Q}_2$, we generate two additional versions using GPT-4, ensuring a total of five.  

We evaluate all three inconsistency types (\topicshort, \randomnessinconsistency, and \paraphraseinconsistency) on MATH dataset, with lower scores indicating better consistency.

\begin{figure}[t]   
    \centering
    \includegraphics[width=0.7\textwidth]{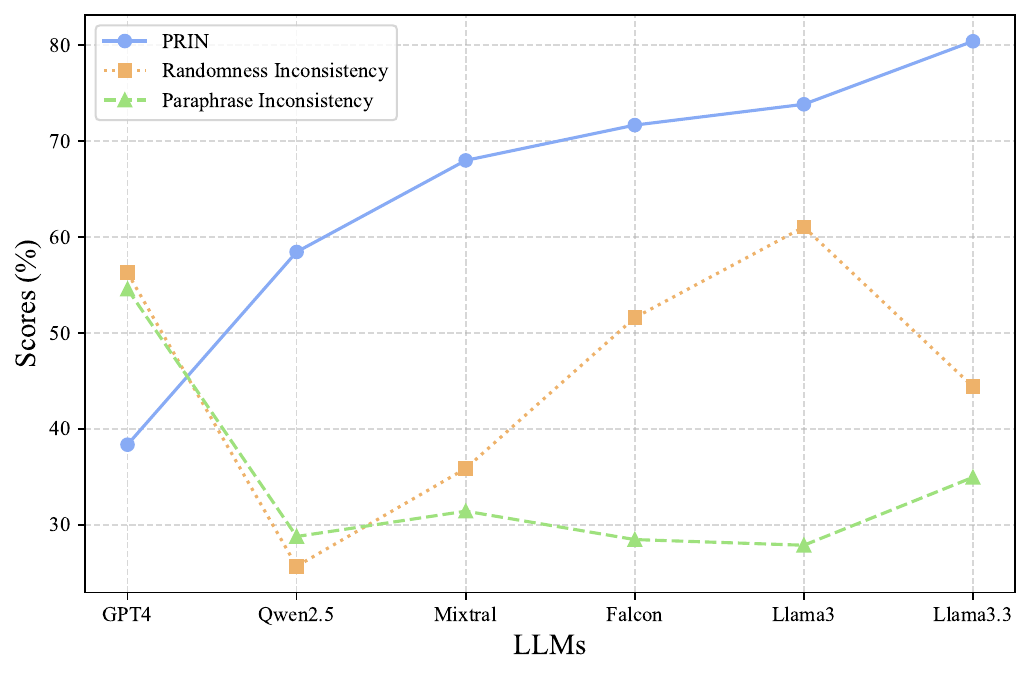}
    \caption{ Scores of \topicshort, \randomnessinconsistency~, and \paraphraseinconsistency~for LLMs.} 
    \label{fig:PRIN_Random_Paraphrase}
\end{figure}

\paragraph{Results.} Figure~\ref{fig:PRIN_Random_Paraphrase} ranks the LLMs based on their \topicshort~scores and additionally presents their \randomnessinconsistency~and \paraphraseinconsistency. This analysis aims to address two key questions: 

\textbf{(i) Why does GPT-4 exhibit higher \randomnessinconsistency~and \paraphraseinconsistency~than most open-source models?} Through error analysis, we observed that GPT-4 tends to follow instructions more faithfully, often formatting its answers using user-specified patterns such as quotation marks, brackets, or colons, even when options are not provided. In contrast, open-source models, regardless of size, frequently fail to follow these formatting instructions and sometimes even terminate generation prematurely without producing valid answers. As a result, when we extract answer spans from these models, we often obtain empty outputs. Therefore, the lower \randomnessinconsistency~and \paraphraseinconsistency~of these models are not due to genuine consistency, but rather stem from consistently producing invalid or incomplete outputs. Importantly, our \topicshort~metric still penalizes such cases when both \direct~and \inverse~outputs are empty, maintaining its diagnostic reliability.

\textbf{(ii) Why does Llama3.3 show lower \randomnessinconsistency~but higher \paraphraseinconsistency~compared to Llama3?} We hypothesize that Llama3.3 has been tuned to behave more deterministically, which mitigates its randomness-driven inconsistency. However, Llama3’s poor instruction-following capability prevents its paraphrase inconsistency from being fully revealed, as it often fails to produce meaningful outputs regardless of paraphrasing. In contrast, Llama3.3 generates more valid outputs due to better instruction-following, thereby exposing its paraphrase inconsistency more clearly.

\begin{table}[t]
    \setlength{\tabcolsep}{4pt}
    \centering
    \begin{tabular}{l|rrr|rrr|rrr}
         & \multicolumn{3}{c|}{MATH} & \multicolumn{3}{c|}{MathQA} & \multicolumn{3}{c}{EquInfer}\\
         &\multicolumn{1}{c}{GPT4} & \multicolumn{1}{c}{GPT-4o} & \multicolumn{1}{c|}{Llama3} &\multicolumn{1}{c}{GPT4} & \multicolumn{1}{c}{GPT-4o} & \multicolumn{1}{c|}{Llama3}
         & \multicolumn{1}{c}{GPT4} & \multicolumn{1}{c}{GPT-4o} & \multicolumn{1}{c}{Llama3} \\\hline
        CoT& 47.58 & 50.67 & 21.55 & 72.57 & 82.73 & 39.03 & 34.81 & 31.27 & 12.60\\
        Self-Consist. & 55.14 & 54.72 & \textbf{26.72} & 79.50 & 85.33 & 42.58& 36.42& 33.94 & \textbf{16.59}\\        
        \topicshort & \textbf{56.44} & \textbf{56.82} & \underline{25.98} & \textbf{82.04} & \textbf{86.63} & \textbf{42.98}& \textbf{37.37}& \textbf{34.51} & \underline{16.02}
    \end{tabular}
    \caption{Comparing \topicshort~with CoT and Self-consistency in promoting LLM performance.}
    \label{tab:promoteperformance}
\end{table}

\subsection{\texorpdfstring{$\mathcal{Q}_5$}{Q5}: How effective can \topicshort~be leveraged to enhance task performance?}

\paragraph{Our Approach:}  
Beyond analyzing \topicshort~as an undesired LLM behavior, we explore how the \direct~and \inverse~can synergize to improve response accuracy. Intuitively, if both \direct~and \inverse~agree that an answer is correct, its correctness probability increases. Based on this insight, our approach selects answers only when both mechanisms indicate correctness.

\paragraph{Setup:}  
Since achieving state-of-the-art performance is not the focus of this study, we conduct a lightweight comparison against widely used prompting strategies, including \textbf{Chain-of-Thought (CoT)} \citep{DBLP:conf/nips/Wei0SBIXCLZ22} and \textbf{Self-Consistency} \citep{DBLP0002WSLCNCZ23}. In addition to GPT-4, we include GPT-4o, a top-performing LLM, to strengthen our hypothesis, as these models are more widely deployed in real-world applications.

\begin{figure}[ht] 
    \centering
    \includegraphics[width=0.8\textwidth]{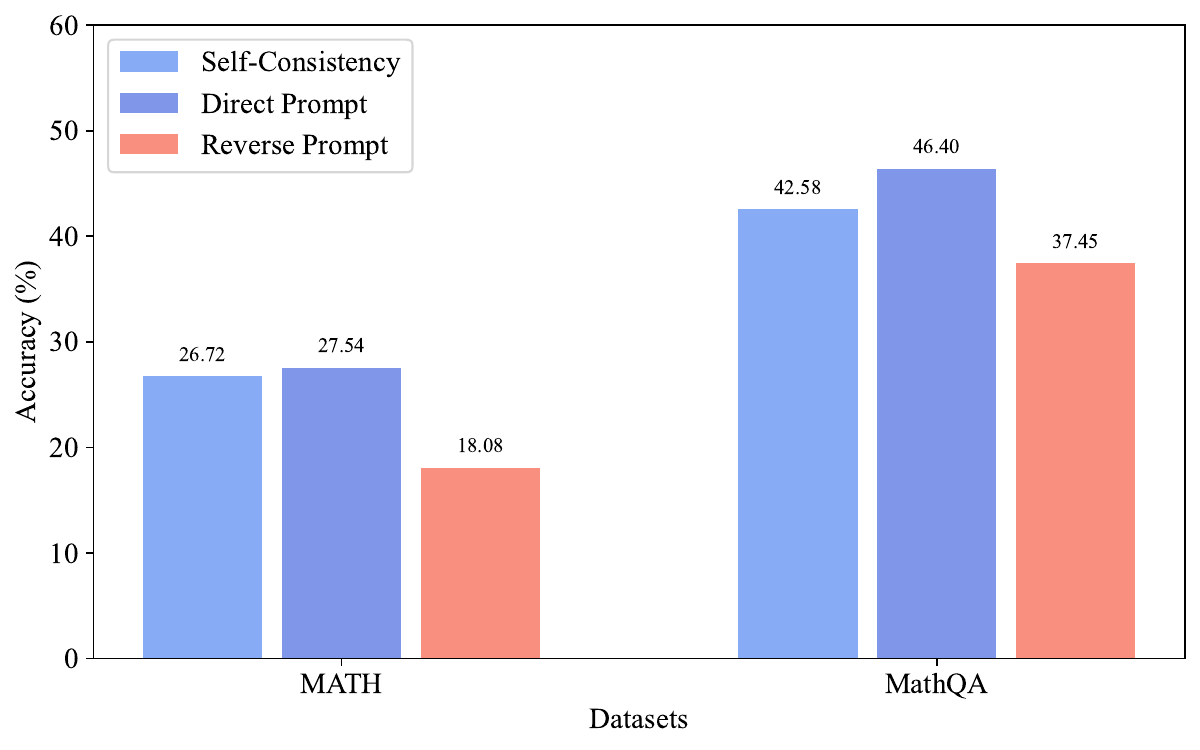}
    \caption{LLMs performance on MATH and MathQA dataset.} 
    \label{fig:performance_comparision}
    \label{fig:reverseprompt}
\end{figure}

\paragraph{Results:} 
Table \ref{tab:promoteperformance} lists the results of \topicshort, CoT and Self-Consistency. We found two quick takeaways:
\begin{itemize}
    \item \topicshort~is very effective to promote top-performing LLM performance, e.g., GPT4 and GPT4o. 
    \item If the LLM is in general weak, \topicshort~do not help (compared with Self-Consistency baseline).
\end{itemize}
We attribute this to the fact that weaker LLMs often struggle with instruction-following, especially when handling negation. To better understand this phenomenon, we break down \topicshort~to further analyze the behavior of \direct~and \inverse~separately. Figure~\ref{fig:reverseprompt} illustrates how Llama3, a representative weaker model, performs when facing Self-Consistency, \direct~and \inverse. Interestingly, Llama3 performs reasonably well when using Self-Consistency or \direct, but its performance drops drastically when only the \inverse~is used. This suggests that weaker models like Llama3 find \inverse~particularly challenging, likely due to difficulties in processing negations, which explains why \topicshort~fails to improve their performance.

\begin{wrapfigure}[12]{r}{0.55\textwidth}
  \begin{center}
\vspace{-15pt}
    \begin{tikzpicture}
        \begin{axis}[
            width=7.5cm, height=4.5cm,  
            xlabel={Number of Options}, 
            ylabel={PRIN},  
            xtick={2,3,4,5},  
            ymin=20, ymax=35,  
            grid=major,  
            no markers,  
            every axis plot post/.append style={thick, mark=o, mark size=3pt, color=blue} 
        ]
        \addplot coordinates {(2,22.46) (3,28.47) (4,30.76) (5,29.96)};
        \end{axis}
    \end{tikzpicture}
\end{center}
\vspace{-15pt}
  \caption{\topicshort~score vs. \#option}
  \label{fig:inconsistency_per_options}
\end{wrapfigure}
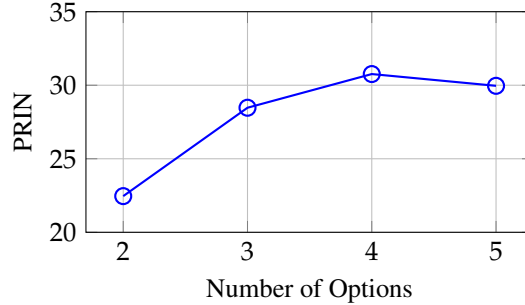

\subsection{\texorpdfstring{$\mathcal{Q}_6$}{Q6}: How does \topicshort~vary with different sizes of options?}

\paragraph{Setup.}
For this experiment, MATH task was given to the GPT4 to derive multiple CoT answers via multiple trials. The 5K problems of MATH were distributed by 4 groups randomly, and each group contains 2,3,4,and 5 distinct answer options. We report \topicshort~for GPT4 for this question.

\paragraph{Results.} Figure \ref{fig:inconsistency_per_options} examines the impact of the number of answer options on \topicshort. Increasing the number of options tends to raise \topicshort~scores, yet models still function within acceptable error limits. This trend underscores the added complexity introduced by a greater variety of options and highlights areas where ongoing algorithm and model improvements are necessary. These findings emphasize the challenges faced by language models in complex decision matrices and open promising avenues for future enhancements in AI development and deployment.

\section{Conclusion}
This study provides a comprehensive analysis of \topic~in LLMs, using diverse tasks and models to explore underlying challenges and potential solutions. By addressing six key questions, our findings stress the importance of integrating reasoning paths and adapting model architectures to optimize performance and reliability. As AI models become increasingly integral across domains, our research underlines the necessity of embedding \topicshort~as a foundational element in model development, ensuring their applicability across diverse, challenging scenarios.

\section*{Acknowledgement}
We would like to sincerely appreciate the anonymous reviewers from OpenReview for their thoughtful insights and constructive suggestions. We are especially grateful to Professor Lili Mou from the University of Alberta for his valuable comments and for posing insightful questions that helped broaden our perspectives. We also deeply appreciate Ibraheem Moosa, Renze Lou, Zhuoyang Zou, Hongchao Fang, and Arshan Dalili for their helpful feedback and suggestions, which played an important role in refining and polishing the final version of this paper.

\bibliography{colm2025_conference}

\begin{thebibliography}{26}
\providecommand{\natexlab}[1]{#1}
\providecommand{\url}[1]{\texttt{#1}}
\expandafter\ifx\csname urlstyle\endcsname\relax
  \providecommand{\doi}[1]{doi: #1}\else
  \providecommand{\doi}{doi: \begingroup \urlstyle{rm}\Url}\fi

\bibitem[Almazrouei et~al.(2023)Almazrouei, Alobeidli, Alshamsi, Cappelli, Cojocaru, Debbah, Étienne Goffinet, Hesslow, Launay, Malartic, Mazzotta, Noune, Pannier, and Penedo]{almazrouei2023falconseriesopenlanguage}
Ebtesam Almazrouei, Hamza Alobeidli, Abdulaziz Alshamsi, Alessandro Cappelli, Ruxandra Cojocaru, Mérouane Debbah, Étienne Goffinet, Daniel Hesslow, Julien Launay, Quentin Malartic, Daniele Mazzotta, Badreddine Noune, Baptiste Pannier, and Guilherme Penedo.
\newblock The falcon series of open language models, 2023.
\newblock URL \url{https://arxiv.org/abs/2311.16867}.

\bibitem[Amini et~al.(2019)Amini, Gabriel, Lin, Koncel-Kedziorski, Choi, and Hajishirzi]{amini2019mathqainterpretablemathword}
Aida Amini, Saadia Gabriel, Peter Lin, Rik Koncel-Kedziorski, Yejin Choi, and Hannaneh Hajishirzi.
\newblock Mathqa: Towards interpretable math word problem solving with operation-based formalisms, 2019.
\newblock URL \url{https://arxiv.org/abs/1905.13319}.

\bibitem[Atil et~al.(2024)Atil, Chittams, Fu, Ture, Xu, and Baldwin]{DBLP04667}
Berk Atil, Alexa Chittams, Liseng Fu, Ferhan Ture, Lixinyu Xu, and Breck Baldwin.
\newblock {LLM} stability: {A} detailed analysis with some surprises.
\newblock \emph{CoRR}, abs/2408.04667, 2024.
\newblock \doi{10.48550/ARXIV.2408.04667}.
\newblock URL \url{https://doi.org/10.48550/arXiv.2408.04667}.

\bibitem[Bubeck et~al.(2023)Bubeck, Chandrasekaran, Eldan, Gehrke, Horvitz, Kamar, Lee, Lee, Li, Lundberg, Nori, Palangi, Ribeiro, and Zhang]{bubeck2023sparksartificialgeneralintelligence}
Sébastien Bubeck, Varun Chandrasekaran, Ronen Eldan, Johannes Gehrke, Eric Horvitz, Ece Kamar, Peter Lee, Yin~Tat Lee, Yuanzhi Li, Scott Lundberg, Harsha Nori, Hamid Palangi, Marco~Tulio Ribeiro, and Yi~Zhang.
\newblock Sparks of artificial general intelligence: Early experiments with gpt-4, 2023.
\newblock URL \url{https://arxiv.org/abs/2303.12712}.

\bibitem[Elazar et~al.(2021)Elazar, Kassner, Ravfogel, Ravichander, Hovy, Schütze, and Goldberg]{elazar2021measuringimprovingconsistencypretrained}
Yanai Elazar, Nora Kassner, Shauli Ravfogel, Abhilasha Ravichander, Eduard Hovy, Hinrich Schütze, and Yoav Goldberg.
\newblock Measuring and improving consistency in pretrained language models, 2021.
\newblock URL \url{https://arxiv.org/abs/2102.01017}.

\bibitem[Errica et~al.(2024)Errica, Siracusano, Sanvito, and Bifulco]{DBLP12334}
Federico Errica, Giuseppe Siracusano, Davide Sanvito, and Roberto Bifulco.
\newblock What did {I} do wrong? quantifying llms' sensitivity and consistency to prompt engineering.
\newblock \emph{CoRR}, abs/2406.12334, 2024.
\newblock \doi{10.48550/ARXIV.2406.12334}.
\newblock URL \url{https://doi.org/10.48550/arXiv.2406.12334}.

\bibitem[Ghazarian et~al.(2024)Ghazarian, Zou, Shah, Peng, Beniwal, Potts, and Sadagopan]{ghazarian2024assessment}
Sarik Ghazarian, Yidong Zou, Swair Shah, Nanyun Peng, Anurag Beniwal, Christopher Potts, and Narayanan Sadagopan.
\newblock Assessment and mitigation of inconsistencies in llm-based evaluations.
\newblock 2024.

\bibitem[Gu et~al.(2023)Gu, Zhao, Xu, Nie, Mei, and Yin]{gu2023robustnesslearningtaskinstructions}
Jiasheng Gu, Hongyu Zhao, Hanzi Xu, Liangyu Nie, Hongyuan Mei, and Wenpeng Yin.
\newblock Robustness of learning from task instructions, 2023.
\newblock URL \url{https://arxiv.org/abs/2212.03813}.

\bibitem[Hendrycks et~al.(2021)Hendrycks, Burns, Kadavath, Arora, Basart, Tang, Song, and Steinhardt]{hendrycks2021measuringmathematicalproblemsolving}
Dan Hendrycks, Collin Burns, Saurav Kadavath, Akul Arora, Steven Basart, Eric Tang, Dawn Song, and Jacob Steinhardt.
\newblock Measuring mathematical problem solving with the math dataset, 2021.
\newblock URL \url{https://arxiv.org/abs/2103.03874}.

\bibitem[Jang \& Lukasiewicz(2023)Jang and Lukasiewicz]{jang2023consistencyanalysischatgpt}
Myeongjun~Erik Jang and Thomas Lukasiewicz.
\newblock Consistency analysis of chatgpt, 2023.
\newblock URL \url{https://arxiv.org/abs/2303.06273}.

\bibitem[Jiang et~al.(2024)Jiang, Sablayrolles, Roux, Mensch, Savary, Bamford, Chaplot, de~las Casas, Hanna, Bressand, Lengyel, Bour, Lample, Lavaud, Saulnier, Lachaux, Stock, Subramanian, Yang, Antoniak, Scao, Gervet, Lavril, Wang, Lacroix, and Sayed]{jiang2024mixtralexperts}
Albert~Q. Jiang, Alexandre Sablayrolles, Antoine Roux, Arthur Mensch, Blanche Savary, Chris Bamford, Devendra~Singh Chaplot, Diego de~las Casas, Emma~Bou Hanna, Florian Bressand, Gianna Lengyel, Guillaume Bour, Guillaume Lample, Lélio~Renard Lavaud, Lucile Saulnier, Marie-Anne Lachaux, Pierre Stock, Sandeep Subramanian, Sophia Yang, Szymon Antoniak, Teven~Le Scao, Théophile Gervet, Thibaut Lavril, Thomas Wang, Timothée Lacroix, and William~El Sayed.
\newblock Mixtral of experts, 2024.
\newblock URL \url{https://arxiv.org/abs/2401.04088}.

\bibitem[Lee et~al.(2024)Lee, Son, Kim, Kim, Chung, Adar, and Kim]{DBLPLeeS0KCAK24}
Yoonjoo Lee, Kihoon Son, Tae~Soo Kim, Jisu Kim, John Joon~Young Chung, Eytan Adar, and Juho Kim.
\newblock One vs. many: Comprehending accurate information from multiple erroneous and inconsistent {AI} generations.
\newblock In \emph{The 2024 {ACM} Conference on Fairness, Accountability, and Transparency, FAccT 2024, Rio de Janeiro, Brazil, June 3-6, 2024}, pp.\  2518--2531. {ACM}, 2024.
\newblock \doi{10.1145/3630106.3662681}.
\newblock URL \url{https://doi.org/10.1145/3630106.3662681}.

\bibitem[Liu et~al.(2024)Liu, Yao, Ton, Zhang, Guo, Cheng, Klochkov, Taufiq, and Li]{liu2024trustworthyllmssurveyguideline}
Yang Liu, Yuanshun Yao, Jean-Francois Ton, Xiaoying Zhang, Ruocheng Guo, Hao Cheng, Yegor Klochkov, Muhammad~Faaiz Taufiq, and Hang Li.
\newblock Trustworthy llms: a survey and guideline for evaluating large language models' alignment, 2024.
\newblock URL \url{https://arxiv.org/abs/2308.05374}.

\bibitem[Lou et~al.(2024)Lou, Xu, Wang, Du, Kamoi, Lu, Xie, Sun, Zhang, Ahn, et~al.]{lou2024aaar}
Renze Lou, Hanzi Xu, Sijia Wang, Jiangshu Du, Ryo Kamoi, Xiaoxin Lu, Jian Xie, Yuxuan Sun, Yusen Zhang, Jihyun~Janice Ahn, et~al.
\newblock Aaar-1.0: Assessing ai's potential to assist research.
\newblock \emph{arXiv preprint arXiv:2410.22394}, 2024.

\bibitem[McIlroy{-}Young et~al.(2024)McIlroy{-}Young, Brown, Olson, Zhang, and Dwork]{DBLPYoungBO24}
Reid McIlroy{-}Young, Katrina Brown, Conlan Olson, Linjun Zhang, and Cynthia Dwork.
\newblock Order-independence without fine tuning.
\newblock In Amir Globersons, Lester Mackey, Danielle Belgrave, Angela Fan, Ulrich Paquet, Jakub~M. Tomczak, and Cheng Zhang (eds.), \emph{Advances in Neural Information Processing Systems 38: Annual Conference on Neural Information Processing Systems 2024, NeurIPS 2024, Vancouver, BC, Canada, December 10 - 15, 2024}, 2024.
\newblock URL \url{http://papers.nips.cc/paper\_files/paper/2024/hash/85529bc995777a74072ef63c05bedd30-Abstract-Conference.html}.

\bibitem[Meta(2024)]{Meta2024Llama3}
Meta.
\newblock Build the future of ai with meta llama 3, 2024.
\newblock URL \url{https://llama.meta.com/llama3/}.
\newblock Accessed: 2024-06-07.

\bibitem[OpenAI(2023)]{openai2023gpt4}
OpenAI.
\newblock Gpt-4 technical report, 2023.

\bibitem[Ravichander et~al.(2022)Ravichander, Gardner, and Marasovi{\'c}]{ravichander2022condaqa}
Abhilasha Ravichander, Matt Gardner, and Ana Marasovi{\'c}.
\newblock Condaqa: A contrastive reading comprehension dataset for reasoning about negation.
\newblock \emph{arXiv preprint arXiv:2211.00295}, 2022.

\bibitem[Team(2024)]{qwen2.5}
Qwen Team.
\newblock Qwen2.5: A party of foundation models, September 2024.
\newblock URL \url{https://qwenlm.github.io/blog/qwen2.5/}.

\bibitem[Wan et~al.(2025)Wan, Wu, Chen, and Li]{wan2025reasoningawareselfconsistencyleveraging}
Guangya Wan, Yuqi Wu, Jie Chen, and Sheng Li.
\newblock Reasoning aware self-consistency: Leveraging reasoning paths for efficient llm sampling, 2025.
\newblock URL \url{https://arxiv.org/abs/2408.17017}.

\bibitem[Wang \& Wang(2025)Wang and Wang]{wang2025assessingconsistencyreproducibilityoutputs}
Julian~Junyan Wang and Victor~Xiaoqi Wang.
\newblock Assessing consistency and reproducibility in the outputs of large language models: Evidence across diverse finance and accounting tasks, 2025.
\newblock URL \url{https://arxiv.org/abs/2503.16974}.

\bibitem[Wang et~al.(2023{\natexlab{a}})Wang, Wei, Schuurmans, Le, Chi, Narang, Chowdhery, and Zhou]{DBLP0002WSLCNCZ23}
Xuezhi Wang, Jason Wei, Dale Schuurmans, Quoc~V. Le, Ed~H. Chi, Sharan Narang, Aakanksha Chowdhery, and Denny Zhou.
\newblock Self-consistency improves chain of thought reasoning in language models.
\newblock In \emph{The Eleventh International Conference on Learning Representations, {ICLR} 2023, Kigali, Rwanda, May 1-5, 2023}. OpenReview.net, 2023{\natexlab{a}}.
\newblock URL \url{https://openreview.net/forum?id=1PL1NIMMrw}.

\bibitem[Wang et~al.(2023{\natexlab{b}})Wang, Wei, Schuurmans, Le, Chi, Narang, Chowdhery, and Zhou]{wangself}
Xuezhi Wang, Jason Wei, Dale Schuurmans, Quoc~V. Le, Ed~H. Chi, Sharan Narang, Aakanksha Chowdhery, and Denny Zhou.
\newblock Self-consistency improves chain of thought reasoning in language models.
\newblock In \emph{The Eleventh International Conference on Learning Representations, {ICLR} 2023, Kigali, Rwanda, May 1-5, 2023}. OpenReview.net, 2023{\natexlab{b}}.
\newblock URL \url{https://openreview.net/forum?id=1PL1NIMMrw}.

\bibitem[Wei et~al.(2022{\natexlab{a}})Wei, Wang, Schuurmans, Bosma, Ichter, Xia, Chi, Le, and Zhou]{DBLP:conf/nips/Wei0SBIXCLZ22}
Jason Wei, Xuezhi Wang, Dale Schuurmans, Maarten Bosma, Brian Ichter, Fei Xia, Ed~H. Chi, Quoc~V. Le, and Denny Zhou.
\newblock Chain-of-thought prompting elicits reasoning in large language models.
\newblock In Sanmi Koyejo, S.~Mohamed, A.~Agarwal, Danielle Belgrave, K.~Cho, and A.~Oh (eds.), \emph{Advances in Neural Information Processing Systems 35: Annual Conference on Neural Information Processing Systems 2022, NeurIPS 2022, New Orleans, LA, USA, November 28 - December 9, 2022}, 2022{\natexlab{a}}.
\newblock URL \url{http://papers.nips.cc/paper\_files/paper/2022/hash/9d5609613524ecf4f15af0f7b31abca4-Abstract-Conference.html}.

\bibitem[Wei et~al.(2022{\natexlab{b}})Wei, Wang, Schuurmans, Bosma, Xia, Chi, Le, Zhou, et~al.]{wei2022chain}
Jason Wei, Xuezhi Wang, Dale Schuurmans, Maarten Bosma, Fei Xia, Ed~Chi, Quoc~V Le, Denny Zhou, et~al.
\newblock Chain-of-thought prompting elicits reasoning in large language models.
\newblock \emph{Advances in neural information processing systems}, 35:\penalty0 24824--24837, 2022{\natexlab{b}}.

\bibitem[Ye et~al.(2023)Ye, Ou, Li, chen, Ma, Yanggong, Wu, Fu, Chen, Wang, and Zhao]{ye2023assessinghiddenrisksllms}
Wentao Ye, Mingfeng Ou, Tianyi Li, Yipeng chen, Xuetao Ma, Yifan Yanggong, Sai Wu, Jie Fu, Gang Chen, Haobo Wang, and Junbo Zhao.
\newblock Assessing hidden risks of llms: An empirical study on robustness, consistency, and credibility, 2023.
\newblock URL \url{https://arxiv.org/abs/2305.10235}.

\end{thebibliography}
\bibliographystyle{colm2025_conference}

\end{document}